\documentclass[10pt,twocolumn,letterpaper]{article}

\usepackage{cvpr}              

\usepackage[dvipsnames]{xcolor}

\definecolor{cvprblue}{rgb}{0.21,0.49,0.74}
\usepackage[pagebackref,breaklinks,colorlinks,allcolors=magenta]{hyperref}
\usepackage{amssymb,bbding,pifont}
\usepackage{tikz}
\usetikzlibrary{positioning}

\definecolor{myblue}{RGB}{66,133,244}
\definecolor{mygreen}{RGB}{51,168,83}
\definecolor{myyellow}{RGB}{251,188,3}
\definecolor{myred}{RGB}{234,67,53}
\definecolor{mygrey}{RGB}{95,99,104}
\definecolor{mypup}{RGB}{153,0,204}

\def\paperID{*****} 
\def\confName{CVPR}
\def\confYear{2026}
\hypersetup{breaklinks,colorlinks}
\title{LoViF 2026 The First Challenge on Holistic Quality Assessment for 4D World Model (PhyScore): Methods and Results}

\author{Wei Luo$^\dagger$\quad Yiting Lu$^\dagger$\quad Xin Li$^\dagger$\quad Haoran Li$^\dagger$\quad Fengbin Guan$^\dagger$ \\
    Chen Gao$^\dagger$\quad Xin Jin$^\dagger$\quad Yong Li$^\dagger$\quad Zhibo Chen$^\dagger$\quad \\
    Sijing Wu\quad Kang Fu\quad Yunhao Li\quad Ziang Xiao\quad Huiyu Duan \\
    Jing Liu\quad Qiang Hu\quad Xiongkuo Min\quad Guangtao Zhai \\
    Manxi Sun\quad Zixuan Guo\quad Yun Li\quad Ziyang Chen\quad Manabu Tsukada \\
    Zhengyang Li\quad Zhenglin Du\quad Yi Wen\quad Licheng Jiao\quad Fang Liu\quad Lingling Li \\
    Yiwen Ren\quad Zhilong Song \\
    Dubing Chen\quad Yucheng Zhou\quad Tianyi Yan\quad Huan Zheng \\
}

\begin{document}
\maketitle
\renewcommand{\thefootnote}{}
\footnotetext{$^\dagger$X. Li (\textcolor{magenta}{xin.li@ustc.edu.cn}), Y. Lu, W. Luo, H. Li, F. Guan, C. Gao,  X. Jin, Y. Li, Z. Chen are challenge organizers.}
\footnotetext{Corresponding author: Zhibo Chen (\textcolor{magenta}{chenzhibo@ustc.edu.cn})}
\footnotetext{Other authors are participants of this challenge.}
\footnotetext{The challenge homepage:~\url{https://www.codabench.org/competitions/13622/}.}
\footnotetext{The LoViF 2026 workshop website:~\url{https://lovif-cvpr2026-workshop.github.io/}.}

\begin{abstract}
This paper reports on the LoViF 2026 PhyScore challenge, a competition on holistic quality assessment of world-model-generated videos across both 2D and 4D generation settings. The challenge is motivated by a central gap in current evaluation practice: perceptual quality alone is insufficient to judge whether generated dynamics are physically plausible, temporally coherent, and consistent with input conditions. Participants are required to build a metric that jointly predicts four dimensions, \emph{i.e.,} Video Quality, Physical Realism, Condition-Video Alignment, and Temporal Consistency. Depart from that, participants also need to localize physical anomaly timestamps for fine-grained diagnosis.

The benchmark dataset contains 1,554 videos generated by seven representative world generative models, organized into three tracks (\texttt{text-2D}, \texttt{image-to-4D}, and \texttt{video-to-4D}) and spanning 26 categories. These categories explicitly cover physics-relevant scenarios, including dynamics, optics, and thermodynamics, together with diverse real-world and creative content. To ensure label reliability, scores and anomaly timestamps are produced through trained human annotation with an additional automated quality-control pass.

Evaluation is based on both score prediction and anomaly localization, with a composite protocol that combines TimeStamp\_IOU and SRCC/PLCC. This report summarizes the challenge design and provides method-level insights from submitted solutions.
\end{abstract}

\section{Introduction}
\label{sec:intro}

World models are evolving rapidly and are being adopted in increasingly broad scenarios, including content creation, virtual simulation, and embodied intelligence systems. A representative family in this line is the predictive generative model, which receives diverse conditions (e.g., text prompts, image prompts, or input videos) and predicts future scenes as outputs, such as video generation \cite{brooks2024video,kling2024,agarwal2025cosmos,yang2024cogvideox,kong2024hunyuanvideo} 3D scene generation\cite{yu2023wonderjourney,wang2024motionctrl,yu2024viewcrafter,chen2025flexworld} and 4D scene generation\cite{zheng2024cami2v,gu2025diffusion,bai2025recammaster,yu2025trajectorycrafter,ex4d}. This trend pushes evaluation beyond appearance quality: a model's understanding of the world should also be reflected in whether the generated evolution obeys physical regularities and remains coherent over time.

Evaluation practice has evolved in parallel with this generative progress. Traditional video quality assessment mainly focuses on perceptual quality in spatial and temporal domains, such as spatial fidelity and temporal stability \cite{wu2022fast,itu_p910_2023,wu2023exploring,guan2025internvqa}. With the rise of AIGC, quality assessment has further incorporated condition consistency, measuring whether generated content aligns with prompts or other input conditions \cite{hessel2021clipscore,lu2024kvq,lu2024aigcvqa,zhang2025q}. Moving to world models, evaluation is shifting toward broader dimensions, including physical plausibility, temporal-causal coherence, assessment of generated 3D/4D content quality, and localized diagnosis of failure moments \cite{huang2024vbench,vbench2025, meng2024phygenbench,duan2025worldscore,lu20254dworldbench,shang2026worldarena}. This progression motivates a holistic evaluation protocol beyond appearance-only scoring.

Consequently, assessing world-model outputs cannot rely on visual fidelity alone. In realistic deployments, generated results are expected to be physically plausible, temporally coherent, and consistent with the provided conditions. Yet this remains difficult in practice: conventional perceptual metrics often miss subtle but critical physical violations (e.g., impossible interactions or inconsistent forces), and purely global scores provide limited diagnostic value when developers need to know where and when such violations occur.

To address this gap, we introduce the LoViF 2026 PhyScore challenge, a benchmark and competition on holistic quality assessment for 4D world model generated videos. Participants are required to design metric and reward models that predict four complementary dimensions: \textbf{VideoQuality} (visual fidelity), \textbf{PhysicalRealism} (physics adherence), \textbf{ConditionVideoAlignment} (consistency with input conditions), and \textbf{Consistency} (spatial-temporal coherence). In addition to score prediction, participants must identify \textbf{physical anomaly timestamps}, localizing time ranges where physical violations occur to enable fine-grained diagnosis.

The challenge dataset is curated to cover heterogeneous generation settings and includes \textbf{1,554} videos generated by \textbf{7} representative world/generative models. It spans three input modalities and corresponding tracks (\texttt{text-2D}, \texttt{image-to-4D}, and \texttt{video-to-4D}), and covers \textbf{26} categories ranging from real-world to creative scenarios. Importantly, physical categories explicitly include principles such as dynamics, optics, and thermodynamics, making the benchmark suitable for evaluating both perceptual quality and physics-related reasoning.

The evaluation protocol emphasizes not only predictive accuracy on the four quality dimensions, but also localization accuracy of anomaly timestamps. The final ranking combines correlation-based score quality and anomaly localization quality (IOU/SRCC/PLCC), encouraging methods that are both reliable and diagnostically informative. To ensure scientific rigor, reproducibility is mandatory for official ranking: top solutions must provide executable code and trained model weights (or equivalent artifacts) for verification.

This report summarizes the challenge setting, dataset construction, evaluation protocol, and performance of top-ranked solutions. We further analyze method trends and trade-offs observed on the leaderboard, highlighting practical insights for building physics-aware and generalizable assessment models for generative video systems.

This challenge is held with the LoViF Workshop~\footnote{\url{https://lovif-cvpr2026-workshop.github.io/}}, which contains a series of challenges: real-world all-in-one image restoration~\cite{lovif2026realir}, efficient VLM for multimodal creative quality scoring~\cite{lovif2026MQualityScoring}, weather removal in videos~\cite{lovif2026WeatherRemoval}, holistic quality assessment for the 4D world model, and human-oriented semantic image quality assessment~\cite{lovif2026SeIQA}.

\section{Challenge}
\label{sec:challenge}
The LoViF 2026 PhyScore challenge aims to advance holistic evaluation of videos generated by world models, including both 2D and 4D generative approaches. Participants are required to submit metric and reward models that can assess generated videos across four dimensions---\textbf{VideoQuality}, \textbf{PhysicalRealism}, \textbf{ConditionVideoAlignment}, and \textbf{Consistency}---and to localize \textbf{physical anomaly timestamps} where violations occur.

\subsection{Datasets}
The dataset is curated by the organizing committee and contains \textbf{1,554} generated videos in total. These videos are produced by \textbf{7 representative world generative models}, including CamI2V\cite{zheng2024cami2v}, Diffusion\_as\_Shader\cite{gu2025diffusion}, ex4d\cite{ex4d}, Recammaster\cite{bai2025recammaster}, TrajctoryCrafter\cite{yu2025trajectorycrafter}, HunyuanVideo\cite{kong2024hunyuanvideo} and others.

To comprehensively evaluate metric and reward models under heterogeneous generation settings, the dataset includes three input modalities: \textbf{text}, \textbf{image}, and \textbf{video}. Correspondingly, samples are organized into three model-type tracks: \texttt{text-2D}, \texttt{image-to-4D}, and \texttt{video-to-4D}.

\begin{figure*}[htbp]
    \centering
    \includegraphics[width=1.0\linewidth]{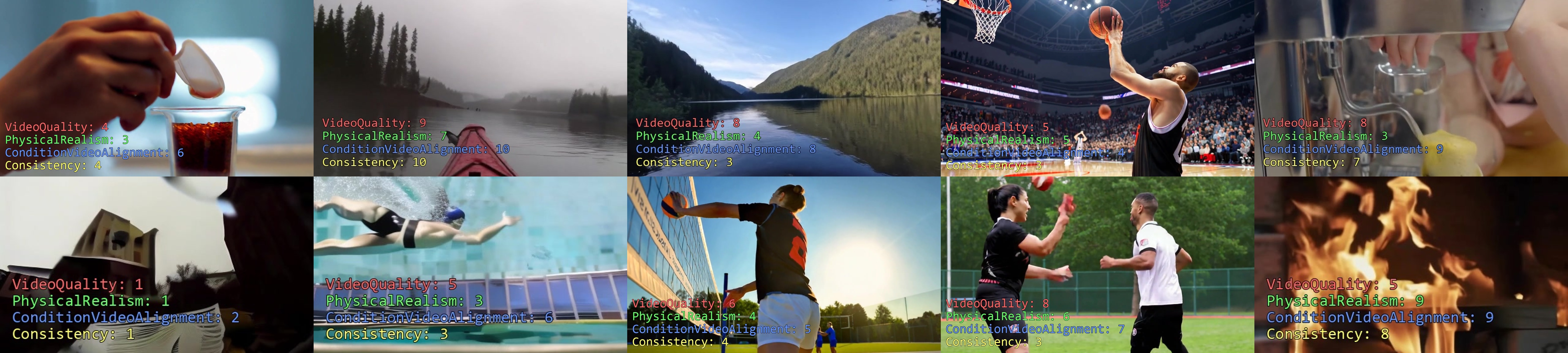}
    \caption{Thumbnails and multidimensional scores of 10 generated videos in the training set: example 1}
    \label{fig:dataset1}
\end{figure*}

\begin{figure*}[htbp]
    \centering
    \includegraphics[width=1.0\linewidth]{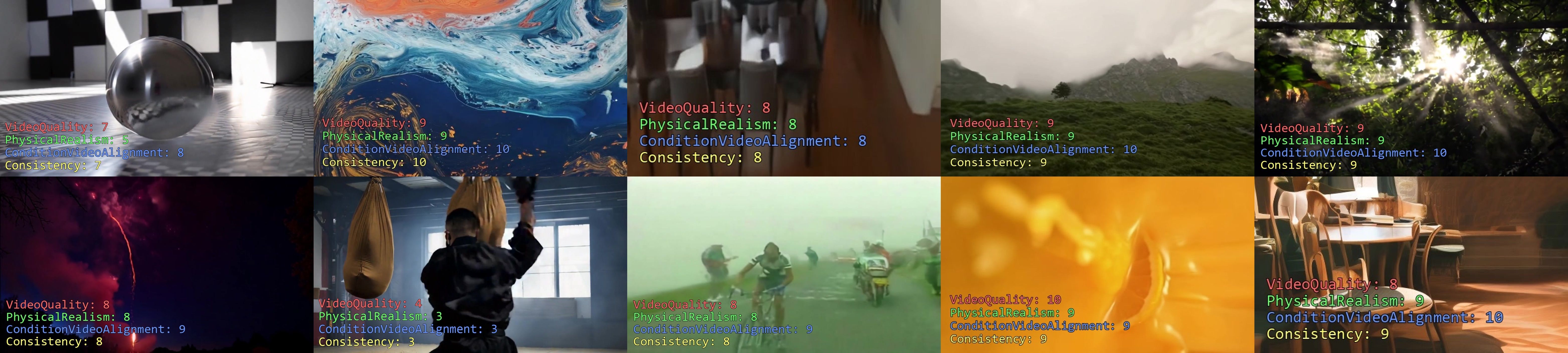}
    \caption{Thumbnails and multidimensional scores of 10 generated videos in the training set: example 2}
    \label{fig:dataset2}
\end{figure*}

\begin{figure*}[t]
    \centering
    \includegraphics[width=1.0\linewidth]{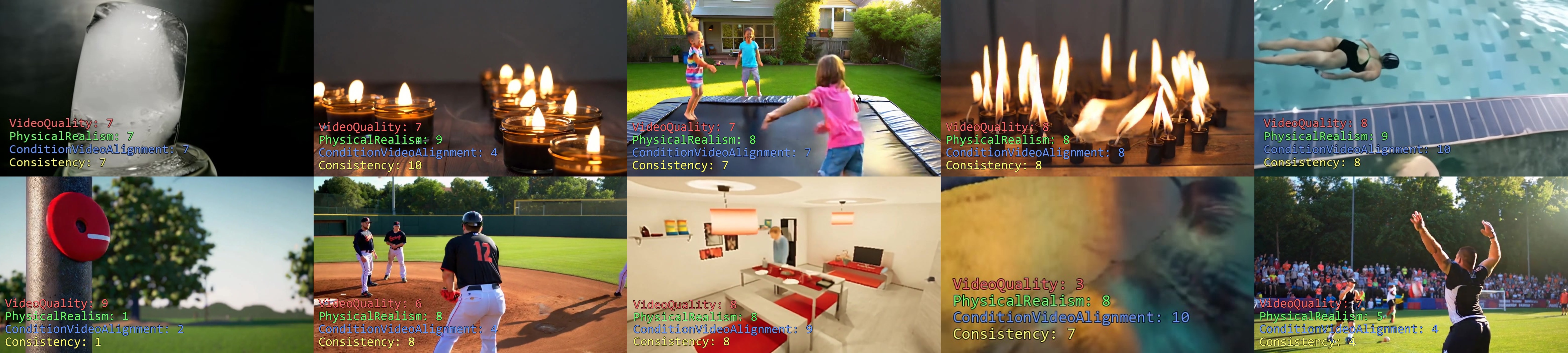}
    \caption{Thumbnails and multidimensional scores of 10 generated videos in the training set: example 3}
    \label{fig:dataset3}
\end{figure*}

The content spans \textbf{26 categories}, covering both physical and non-physical scenarios. In particular, the physical categories explicitly include \textbf{dynamics}, \textbf{optics}, and \textbf{thermodynamics}, each with both high-motion and low-motion cases. The remaining categories further broaden scene diversity with sports, games, movies, cartoons, egocentric views, and challenging environmental conditions such as rainy, foggy, and night scenes. Some thumbnails and multidimensional scores of  the dataset are shown in Fig\ref{fig:dataset1}, \ref{fig:dataset2}, \ref{fig:dataset3}

\textbf{Subjective Annotation Details.} 
To ensure a high-quality benchmark, the ground-truth scores were collected through a rigorous subjective annotation process. We employed a specialized web-based annotation system. Four professional annotators were carefully trained with detailed scoring instructions before the task. They were asked to rate each video on a scale of 1 to 10 across the four dimensions: Video Quality, Physical Realism, Condition-Video Alignment, and Spatial-Temporal Consistency. Crucially, when evaluating Physical Realism, annotators were required to explicitly identify the specific time ranges (timestamps) where any physical violations occurred (e.g., "00:02-00:05"). 

After the initial human annotation phase, we performed an automated quality control pass. Specifically, we used the Gemini model to review the annotations, focusing on cases where there were inconsistencies between the given Physical Realism score and the annotated anomaly timestamps. This refinement step ensures that the temporal localization of physical violations strictly aligns with the corresponding holistic realism score, producing highly reliable labels for the challenge.

The released data is split into:
\begin{itemize}
    \item \textbf{Train}: 863 videos with ground-truth annotations for the four dimension scores and anomaly timestamp labels.
    \item \textbf{Validation}: 298 videos (inputs only) for development and tuning.
    \item \textbf{Test}: 393 videos (inputs only) reserved for final evaluation.
\end{itemize}

\subsection{Evaluation Protocol}
For each test video, participants submit predicted scores for the four dimensions and an anomaly timestamp range (or \texttt{None} if no violation is present). The anomaly timestamp is used to compute the physical anomaly detection IoU (TimeStamp\_IOU on the leaderboard). The overall ranking is based on a composite score that balances correlation with the ground truth scores and the anomaly localization quality:
\begin{equation}
    \mathrm{Final\_Score} = 0.2\times \mathrm{IOU} + 0.4\times \mathrm{SRCC} + 0.4\times \mathrm{PLCC},
    \label{eq:finalscore}
\end{equation}
where SRCC and PLCC are computed on the four dimension predictions (averaged across the four dimensions), and IOU is computed from the predicted anomaly timestamp ranges.

\subsection{Challenge Phases}
There are two phases in this challenge, \ie, the development and testing phases. The details are as follows.

\noindent\textbf{Development Phase:}
In the development phase, the training set (863 videos with ground truth) and the validation set (298 videos with inputs only) are provided. Participants can iteratively submit predictions to the validation server to tune their models. To encourage scientific rigor, participants are asked to prepare runnable code and model artifacts for later reproducibility verification.

\begin{table*}[tp]
    \centering
    \caption{Quantitative results on the PhyScore testing leaderboard (CodaBench). The best and second best values per column are highlighted in \textcolor{red}{red} and \textcolor{blue}{blue}, respectively. For \textbf{ConditionVideoAlignment} (sorted ascending), \textbf{Runtime}, and \textbf{ExtraData}, lower is better as defined on the leaderboard. TimeStamp\_IOU measures anomaly timestamp localization quality. The ranking index matches the numeric prefix of the corresponding team folder in this project.}
    \resizebox{\textwidth}{!}{
    \begin{tabular}{c|c|cccccc|ccc}
    \toprule
    Rank & Team & Final Score $\uparrow$ & VideoQuality $\uparrow$ & PhysicalRealism $\uparrow$ & ConditionVideoAlignment $\downarrow$ & Consistency $\uparrow$ & TimeStamp\_IOU $\uparrow$ & Runtime [s] $\downarrow$ & CPU[1]/GPU[0] & ExtraData[1]/No[0] $\downarrow$\\ \midrule
    1 & SJTU\_MM & \textcolor{red}{0.532} & 0.457 & \textcolor{red}{0.497} & 0.582 & \textcolor{red}{0.519} & \textcolor{red}{0.607} & \textcolor{red}{12.0} & 0 & \textcolor{blue}{1} \\
    2 & IHNI & \textcolor{blue}{0.525} & \textcolor{blue}{0.467} & \textcolor{blue}{0.489} & 0.562 & \textcolor{blue}{0.501} & \textcolor{blue}{0.605} & 15.0 & 0 & \textcolor{red}{0} \\
    3 & WDL & 0.486 & 0.429 & 0.447 & \textcolor{blue}{0.517} & 0.457 & 0.580 & \textcolor{blue}{12.0} & 0 & 1 \\
    4 & zhilong & 0.484 & \textcolor{red}{0.478} & 0.417 & 0.528 & 0.460 & 0.536 & 12.0 & 0 & 1 \\
    5 & DYTH & 0.399 & 0.362 & 0.305 & \textcolor{red}{0.461} & 0.329 & 0.538 & 60.0 & 0 & 1 \\
    \bottomrule
    \end{tabular}}
    \label{tab:results}
\end{table*}

\noindent\textbf{Testing Phases:}
In the testing phase, the test set (393 videos with inputs only) is released for final evaluation. For official ranking and eligibility, participants must submit a factsheet and reproducible code/executable (e.g., Docker) for verification, consistent with the challenge rules on CodaBench. The final ranking is computed with Eq.~\ref{eq:finalscore}.
\section{Challenge Results}
We have summarized the challenge results in Table~\ref{tab:results}. 
Overall, the leaderboard reflects diverse design choices for holistic evaluation. The top-ranked submission (SJTU-MM) achieves the best overall Final Score, driven by strong performance on \textbf{PhysicalRealism}, \textbf{Consistency}, and \textbf{TimeStamp\_IOU}, while maintaining an efficient runtime. IHNI closely follows with consistently strong correlations across dimensions and competitive anomaly localization. Notably, different teams exhibit different strengths: zhilong achieves the best \textbf{VideoQuality} score, while DYTH obtains the lowest (best) \textbf{ConditionVideoAlignment} value as defined by the leaderboard sorting rule.

These results highlight key trade-offs in metric learning for world-model videos. Methods that emphasize physics-aware representations tend to benefit \textbf{PhysicalRealism} and anomaly timestamp localization, whereas specialized perceptual modeling can improve \textbf{VideoQuality}. In addition, runtime and extra-data usage further differentiate practical deployments, suggesting that effective holistic evaluators must balance accuracy, generalization, and efficiency.

\subsection{Experimental Analysis}
Drawing inspiration from recent short-form UGC video quality assessment challenges, we summarize several key findings and common practices from the submitted methods that lead to superior performance in holistic quality assessment for 4D world models.

\textbf{1. Architectures and Feature Extractors:}
A majority of the participating teams utilize multi-branch architectures to capture different dimensions of video quality. Spatial features are frequently extracted using powerful backbones such as ConvNeXt or Swin Transformer, while temporal and motion distortions are captured via SlowFast networks or VideoMAE. Furthermore, the integration of Large Multi-Modality Models (LMMs) and foundational visual encoders has proven highly effective in aligning visual features with semantic and quality-aware representations, which is crucial for evaluating \textbf{ConditionVideoAlignment} and \textbf{VideoQuality}.

\textbf{2. Physics-aware and Anomaly Localization Strategies:}
Unlike traditional UGC video quality assessment, evaluating videos generated by 4D world models requires specialized modeling for physical realism and spatio-temporal consistency. Methods that explicitly incorporate physics-aware representations or optical flow modules tend to achieve better anomaly timestamp localization (\textbf{TimeStamp\_IOU}) and higher \textbf{PhysicalRealism} scores. This indicates that tracking temporal dynamics and physical constraints is indispensable for holistic world-model evaluation.

\textbf{3. Ensemble Methods and Training Strategies:}
Consistent with the findings in existing VQA challenges, ensemble strategies are widely adopted by the top teams to boost prediction accuracy and stability. Model ensembles across different network architectures or fusing model weights from different training epochs significantly enhance the robustness of the final predictions. Additionally, training tricks such as Exponential Moving Average (EMA), rank-aware loss optimization, and advanced data augmentation strategies are commonly utilized to prevent overfitting and ensure stable model convergence.

\textbf{4. Extra Data and Pre-training:}
The employment of extra training data and pre-trained weights is prevalent among the leading submissions. Leveraging models pre-trained on large-scale natural or synthetic video datasets provides a strong prior for general video quality assessment. Specifically, utilizing additional datasets helps the models generalize better to the complex and diverse distortions generated by world models, thereby yielding more accurate predictions across the four evaluated dimensions.

\textbf{5. Trade-offs between Accuracy and Efficiency:}
These results highlight key trade-offs in metric learning for world-model videos. Methods that emphasize elaborate spatio-temporal tracking or physics-aware modules tend to benefit \textbf{PhysicalRealism} and anomaly localization, whereas specialized perceptual models directly improve \textbf{VideoQuality}. In addition, runtime and extra-data usage further differentiate practical deployments. As observed in the leaderboard, achieving a higher Final Score often involves increased computational complexity. This suggests that future effective holistic evaluators must balance accuracy, generalization, and computational efficiency for large-scale practical deployments.
\section{Teams and Methods}
\label{sec:teams_and_methods}


\subsection{SJTU\_MM}

\begin{figure*}[t]
    \centering
    \includegraphics[width=1.0\linewidth]{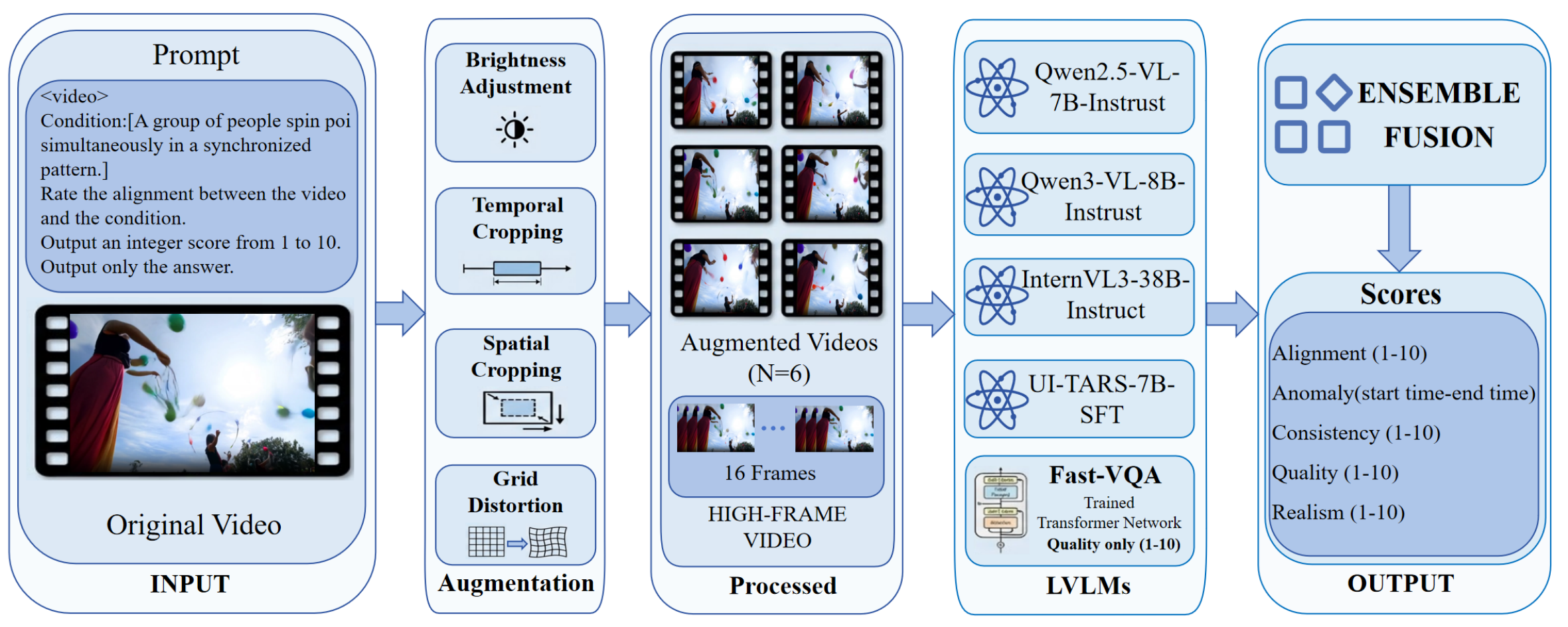}
    \caption{Overview of team SJTU-MM proposed method.}
    \label{fig:SJTU_MM}
\end{figure*}

This team proposed a method mainly utilizes multiple large multi-modal models (LMMs) and fine-tunes them on the training dataset. The whole framework is summarized in Figure \ref{fig:SJTU_MM} and it contains several steps: Data processing and augmentation (step I), Model selection and training (step II), and Score ensemble (step III). 

\noindent\textbf{Method Description.}
For step I, all videos are first uniformly sampled to 8 frames while preserving original resolutions, then expanded with higher-frame sampling (16 frames/video), four augmentations (brightness adjustment, temporal cropping, spatial cropping, and grid distortion), and resolution unification to fixed long-side resolutions (768 and 1440). For step II, the team uses Qwen3-VL-8B \cite{bai2025qwen3}, Qwen2.5-VL-7B \cite{Qwen25VL}, InternVL3-38B \cite{zhu2025internvl3}, and UI-TARS-7B-SFT \cite{qin2025ui} as base LMMs to jointly predict video quality, physical realism, condition--video alignment, consistency, and timestamps. For step III, per-dimension score selection and averaging are applied to produce the final submission.

\noindent\textbf{Training Information.}
The LMMs are fine-tuned with supervised fine-tuning (SFT) and low-rank adaptation (LoRA), optimizing the vision encoder, projector, and language model components. Training is implemented with MS-Swift \cite{zhao2025swift} for 10 epochs. In parallel, the team trains FAST-VQA \cite{wu2022fast} specifically for the video-quality dimension and incorporates its predictions into the final ensemble.

\noindent\textbf{Testing Information.}
During testing, predictions from multiple fine-tuned models are selectively fused for each evaluation dimension and then averaged for final outputs. According to the team factsheet, no extra challenge data is used beyond the official PhyScore setting.

\subsection{IHNI}
\begin{figure}
    \centering
    \includegraphics[width=1.0\linewidth]{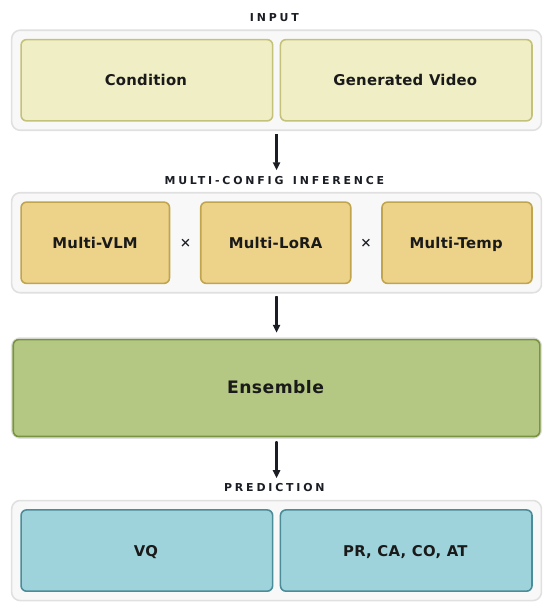}
    \caption{Overview of team INHI proposed method.}
    \label{fig:INHI}
\end{figure}

\noindent\textbf{Method Description.}
IHNI proposes MOSAIC (\emph{Multi-Observer Simulation via Adapter and Inference Configuration}) as Fig\ref{fig:INHI}, which emulates a panel of observers by combining heterogeneous adapter settings and controlled inference temperatures. The method uses two vision-language backbones, VideoScore2 (Qwen2.5-VL-7B based) and Qwen3-VL-8B-Instruct \cite{he2025videoscore2,bai2025qwen3}. For each backbone, the team trains eight LoRA adapters, split into four \emph{4dim} adapters (for physical realism, condition--video alignment, consistency, and anomaly ranges) and four \emph{VQ} adapters (video quality only), resulting in 16 adapters in total. Each adapter is queried at six temperatures ($T \in \{0.1,0.2,0.3,0.4,0.5,0.6\}$), producing 96 prediction variants that provide complementary observer-like behaviors. For score fusion, PR/CA/CO predictions are aggregated by consistency-weighted rank averaging inspired by ITU-T P.910 \cite{itu_p910_2023}, VQ is obtained by averaging VQ-specialist outputs, and anomaly ranges are generated from 4dim consensus with group-wise statistics and majority voting, followed by affine calibration.

\noindent\textbf{Training Information.}
The team trains only on the official PhyScore training set (863 samples), without extra labeled data. LoRA configurations are diversified by rank and learning rate (ranks 64/128/256 with learning rates $1\times10^{-4}$ or $5\times10^{-5}$), with dropout set to 0.05. Training runs for 3 epochs using AdamW (weight decay 0.01), cosine learning-rate decay, and 10\% warmup, under bfloat16 precision. To match GPU memory constraints while keeping optimization stable, the per-step batch size is 1 and gradient accumulation is set to 16 (effective batch size 16). Input design is explicitly task-specific: the 4dim branch emphasizes temporal density (training fps sampled from [2.0, 6.0], inference at fps=4.0, max\_pixels=129{,}600), whereas the VQ branch favors spatial fidelity (training fps sampled from [1.0, 3.0], inference at fps=2.0, max\_pixels=360{,}000).

\noindent\textbf{Testing Information.}
At inference, the full adapter-temperature pool is executed and fused in a task-aware manner: VQ predictions are averaged from VQ-specialist variants, PR/CA/CO are combined through consistency-weighted rank aggregation, and anomaly-related outputs are derived from 4dim consensus with group-wise statistics and majority voting before calibration. The factsheet reports a best test score of 0.525 (submission ID 649121), corresponding to second place, with per-metric values of 0.467 (video quality), 0.489 (physical realism), 0.562 (condition--video alignment), 0.501 (consistency), and 0.605 (timestamp IoU).

\subsection{WDL}

\begin{figure*}[t]
  \centering
  \resizebox{\textwidth}{!}{
    \begin{tikzpicture}[
    node distance=9mm and 10mm,
    box/.style={draw, rounded corners=2pt, align=center, minimum height=8mm, inner sep=3pt, fill=blue!4},
    io/.style={draw, rounded corners=2pt, align=center, minimum height=8mm, inner sep=3pt, fill=green!6},
    head/.style={draw, rounded corners=2pt, align=center, minimum height=8mm, inner sep=3pt, fill=orange!8},
    arrow/.style={-latex, thick}
]

\node[io] (video) {Input Video};
\node[box, right=of video] (sample) {Uniform Frame Sampling\\(4 frames)};
\node[box, right=of sample] (backbone) {Frozen InternVideo2\\(1B-224p-f4)};
\node[box, below=of backbone] (meta) {Metadata Encoder\\One-hot(type, model)};
\node[box, right=of backbone] (concat) {Feature Concatenation\\video + metadata};
\node[head, right=of concat] (head) {QualityHeadV2\\Multi-task MLP};

\node[io, above right=3mm and 12mm of head] (score) {4 Quality Scores\\(1--10)};
\node[io, right=16mm of head] (prob) {Anomaly Probability};
\node[io, below right=3mm and 12mm of head] (ts) {Anomaly Timestamps\\(start, end)};
\node[box, below=of head] (ens) {5-fold CV Ensemble\\(mean over 5 heads)};

\draw[arrow] (video) -- (sample);
\draw[arrow] (sample) -- (backbone);
\draw[arrow] (backbone) -- (concat);
\draw[arrow] (meta) -| (concat);
\draw[arrow] (concat) -- (head);
\draw[arrow] (head) -- (score);
\draw[arrow] (head) -- (prob);
\draw[arrow] (head) -- (ts);
\draw[arrow] (head) -- (ens);

\end{tikzpicture}
  }
  \caption{Overview of the WDL two-stage pipeline and multi-task outputs.}
  \label{fig:WDL}
\end{figure*}
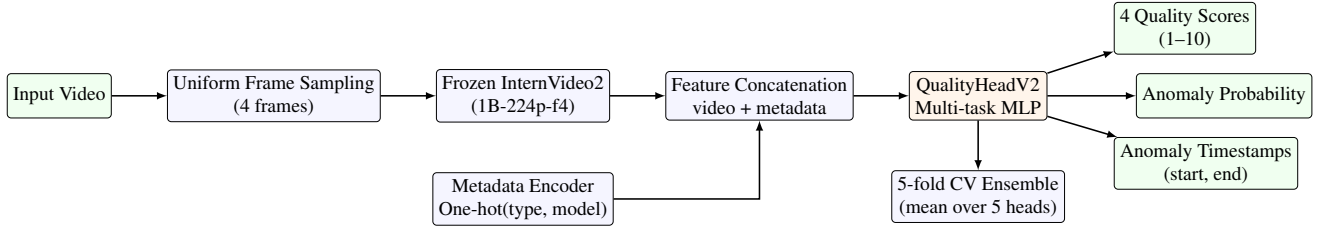

\noindent\textbf{Method Description.}
WDL adopts a two-stage pipeline in Fig~\ref{fig:WDL}. First, a frozen InternVideo2 backbone \cite{wang2022internvideo} (Stage2, 1B-224p-f4) extracts 768-dimensional visual features from 8 uniformly sampled frames. Second, these features are concatenated with one-hot encodings of video \emph{type} and generator \emph{model name}, then fed into a lightweight multi-task head (\emph{QualityHeadV2}, an MLP with LayerNorm and GELU) that contains three branches: 4D score regression, anomaly probability classification, and timestamp regression. The head outputs four quality scores, anomaly probability, and anomaly start/end timestamps.

\noindent\textbf{Training Information.}
Training uses only official challenge data with no extra samples. The total loss is
\[
\mathcal{L}=\mathcal{L}_{\text{score}} + 0.35\,\mathcal{L}_{\text{prob}} + 0.40\,\mathcal{L}_{\text{ts}},
\]
where $\mathcal{L}_{\text{score}}$ is Huber loss on quality scores (VideoQuality weighted by 1.8 and the other three dimensions by 1.0), $\mathcal{L}_{\text{prob}}$ is binary cross-entropy for anomaly probability, and $\mathcal{L}_{\text{ts}}$ is Smooth L1 loss for timestamps. Optimization uses AdamW with learning rate $3\times10^{-4}$ and weight decay $10^{-4}$, batch size 64, up to 220 epochs with early stopping patience 35. The team trains five models in 5-fold cross-validation, and monitors $\text{IoU} - 0.08\times\text{MAE}$ as the early-stopping validation metric.

\noindent\textbf{Testing Information.}
During inference, the five fold models are ensembled by arithmetic averaging of predictions, and no external data is used. The factsheet reports average validation composite score $\approx 0.420$ and feature extraction speed of about 0.39\,s per video. A public test leaderboard score is not provided in the official team factsheet.

\subsection{zhilong}

\noindent\textbf{Method Description.}
Xidian University proposes a type-separated ensemble pipeline that keeps CLIP \cite{radford2021learningclip} frozen and combines deep features with handcrafted statistics. The feature set includes frame-level CLIP embeddings, appearance statistics (e.g., brightness/contrast/saturation/sharpness), temporal descriptors (e.g., frame-difference energy and motion-related cues), and metadata-derived signals. For prediction, the team combines Random Forests \cite{breiman2001randomforest}, Extra Trees \cite{geurts2006extremely}, Ridge regression \cite{hoerl1970ridge}, and histogram-based gradient boosting \cite{friedman2001greedy}, with optional same-type $k$-NN refinement \cite{cover1967nearestneighbor}, plus a dedicated anomaly timestamp branch.

\noindent\textbf{Training Information.}
The pipeline is trained and inferred separately for official data types (\texttt{text-2D}, \texttt{image-4D}, \texttt{video-4D}), which is a key design choice in this submission. The team reports 5-fold type-stratified cross-validation for hyperparameter selection and then retrains selected settings on all official training data. No extra challenge data is used; external resources are limited to pretrained CLIP weights.

\noindent\textbf{Testing Information.}
The factsheet reports a public test score of 0.484 and a local cross-validation score of 0.4936 for the type-separated setting. Inference uses per-type models and an anomaly branch with post-hoc temporal correction based on motion dynamics.

\subsection{DYTH}
\begin{figure}
    \centering
    \includegraphics[width=1.0\linewidth]{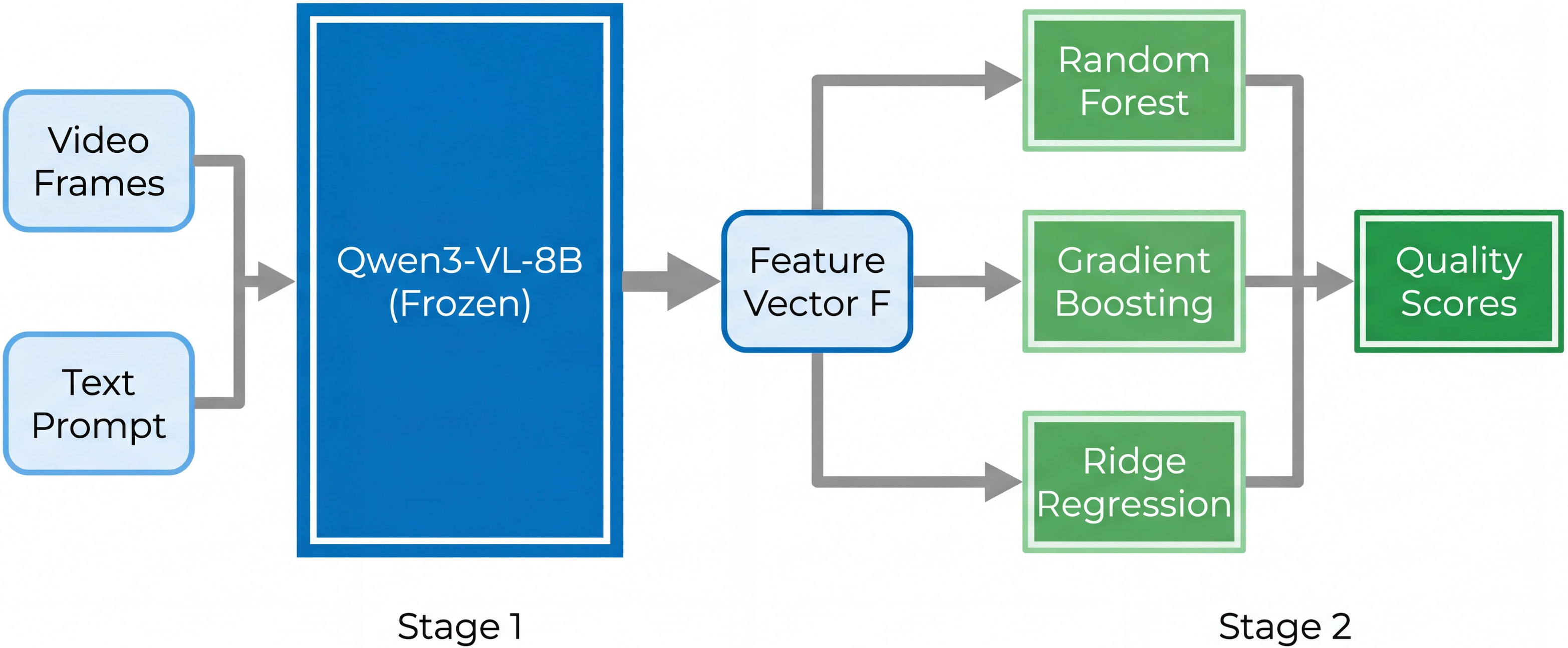}
    \caption{Overview of team DYTH proposed method.}
    \label{fig:DYTH}
\end{figure}

\noindent\textbf{Method Description.}
DYTH proposes a decoupled pipeline where representation learning and regression are separated, shown as Fig~\ref{fig:DYTH}. Qwen3-VL-8B-Instruct \cite{bai2025qwen3} is used as a frozen feature extractor: uniformly sampled keyframes are encoded jointly with text conditions, and penultimate hidden states are used as text-conditioned embeddings. A lightweight heterogeneous ensemble (Random Forest, Gradient Boosting, and Ridge) maps embeddings to five targets (four quality dimensions plus temporal IoU), followed by per-dimension averaging. Within the ensemble, Random Forest captures non-linear feature interactions, Gradient Boosting focuses on residual hard cases, and Ridge stabilizes linear trends to reduce extreme predictions.

\noindent\textbf{Training Information.}
Only official competition data is used, and the VLM backbone remains frozen while lightweight prediction heads are trained. The team reports internal ablations over three system versions: Version 1 gives a baseline FinalScore of 0.3304, Version 2 improves alignment but causes a large PhysicalRealism drop (0.1249), and Version 3 (frozen backbone + heterogeneous ensemble) improves cross-metric balance. Detailed optimizer-level hyperparameters are not provided in the official team factsheet.

\noindent\textbf{Testing Information.}
The final submission follows the same frozen-backbone plus ensemble-head strategy at test time. According to the official team factsheet, the final submission reports 0.3616 (Video Quality), 0.3048 (Physical Realism), 0.4608 (Condition--Video Alignment), 0.3293 (Consistency), and 0.5380 (Temporal IoU), with FinalScore 0.3989.

\appendix

\subsection*{Organizers}
\label{organizers}
\noindent\textit{\textbf{Title:}} LoViF 2026 The First Challenge on Holistic Quality Assessment for 4D World Model (PhyScore)@CVPR2026

\noindent\textit{\textbf{Members:}} 

\noindent Xin Li (\textcolor{magenta}{xin.li@ustc.edu.cn}),

\noindent Yiting Lu (\textcolor{magenta}{luyt31415@mail.ustc.edu.cn}),

\noindent Wei Luo (\textcolor{magenta}{lw21@mail.ustc.edu.cn}),

\noindent Haoran Li (\textcolor{magenta}{lihr@mail.ustc.edu.cn}),

\noindent Fengbin Guan (\textcolor{magenta}{guanfb@mail.ustc.edu.cn}),

\noindent Chen Gao (\textcolor{magenta}{chgao96@tsinghua.edu.cn}), 

\noindent Xin Jin (\textcolor{magenta}{jinxin@eitech.edu.cn}), 

\noindent Yong Li (\textcolor{magenta}{liyong07@tsinghua.edu.cn}), 

\noindent Zhibo Chen (\textcolor{magenta}{chenzhibo@ustc.edu.cn})

\noindent\textit{\textbf{Website:}} \url{https://lovif-cvpr2026-workshop.github.io/}

\noindent\textit{\textbf{Competition page:}} \url{https://www.codabench.org/competitions/13622/}

\subsection*{Participating Teams}

\subsection*{SJTU\_MM}
\noindent\textit{\textbf{Members:}} Sijing Wu, Kang Fu, Yunhao Li, Ziang Xiao, Huiyu Duan, Jing Liu, Qiang Hu, Xiongkuo Min, Guangtao Zhai

\noindent\textit{\textbf{Affiliations:}} Shanghai Jiao Tong University; Tianjin University

\subsection*{IHNI}
\noindent\textit{\textbf{Members:}} Manxi Sun, Zixuan Guo, Yun Li, Ziyang Chen, Manabu Tsukada

\noindent\textit{\textbf{Affiliations:}} Futian Lab; The University of Tokyo

\subsection*{WDL}
\noindent\textit{\textbf{Members:}} Zhengyang Li, Zhenglin Du, Yi Wen, Licheng Jiao, Fang Liu, Lingling Li

\noindent\textit{\textbf{Affiliations:}} Xidian University

\subsection*{zhilong}
\noindent\textit{\textbf{Members:}} Yiwen Ren, Zhilong Song

\noindent\textit{\textbf{Affiliations:}} Xidian University

\subsection*{DYTH}
\noindent\textit{\textbf{Members:}} Dubing Chen, Yucheng Zhou, Tianyi Yan, Huan Zheng

\noindent\textit{\textbf{Affiliations:}} University of Macau

\subsection*{Acknowledgments}
This work was supported by Grants of NSFC 62302246, 62371434, ZJNSFC LQ23F010008, Ningbo 2023Z237 \& 2024Z284 \& 2024Z289 \& 2023CX050011 \& 2025Z038 \& 2025Z059, and supported by High Performance Computing Center at Eastern Institute of Technology and Ningbo Institute of Digital Twin, the Fundamental Research Funds for the Central Universities (No. WK2100250064) and ZGCA Project-C20250302.

{
    \small
    \bibliographystyle{ieeenat_fullname}
    \bibliography{main}

@String(CVPR= {IEEE Conf. Comput. Vis. Pattern Recog.})

@String(AAAI = {AAAI})

@String(CVPR  = {CVPR})

@inproceedings{lovif2026realir,
  title={{LoViF} 2026 Challenge on Real-World All-in-One Image Restoration: Methods and Results},
  author={Chen, Xiang and Li, Hao and Dong, Jiangxin and Pan, Jinshan and Li, Xin and others},
  booktitle={Proceedings of the IEEE/CVF Conference on Computer Vision and Pattern Recognition (CVPR) Workshops},
  year={2026}
}

@inproceedings{lovif2026MQualityScoring,
  title={The 1st {LoViF} Challenge on Efficient VLM for Multimodal Creative Quality Scoring: Methods and Results},
  author={Zhang, Jusheng and Lyu, Qinhan and Ma,  Sizhuo and Cao, Sheng and Wang, Jian and Li, Xin and Wang, Keze and Zheng, Yongsen and Yang, Jing and others},
  booktitle={Proceedings of the IEEE/CVF Conference on Computer Vision and Pattern Recognition (CVPR) Workshops},
  year={2026}
}

@inproceedings{lovif2026WeatherRemoval,
  title={{LoViF} 2026 The First Challenge on Weather Removal in Videos},
  author={Qian, Chenghao and Li, Xin and Jin, Yeying and Sun, Shangquan and others},
  booktitle={Proceedings of the IEEE/CVF Conference on Computer Vision and Pattern Recognition (CVPR) Workshops},
  year={2026}
}

@inproceedings{lovif2026SeIQA,
  title={{LoViF} 2026 The First Challenge on Human-Oriented Semantic Image Quality Assessment: Methods and Results},
  author={Li, Xin and Xu, Daoli	and Luo,  Wei and Xiang, Guoqiang and Li, Haoran and Zhuang, Chengyu and Chen, Zhibo and Guan, Jian and Li, Weipingand others},
  booktitle={Proceedings of the IEEE/CVF Conference on Computer Vision and Pattern Recognition (CVPR) Workshops},
  year={2026}
}

@inproceedings{radford2021learningclip,
  title={Learning transferable visual models from natural language supervision},
  author={Radford, Alec and Kim, Jong Wook and Hallacy, Chris and Ramesh, Aditya and Goh, Gabriel and Agarwal, Sandhini and Sastry, Girish and Askell, Amanda and Mishkin, Pamela and Clark, Jack and others},
  booktitle={International conference on machine learning},
  pages={8748--8763},
  year={2021},
  organization={PmLR}
}

@inproceedings{lu2024aigcvqa,
  title={Aigc-vqa: A holistic perception metric for aigc video quality assessment},
  author={Lu, Yiting and Li, Xin and Li, Bingchen and Yu, Zihao and Guan, Fengbin and Wang, Xinrui and Liao, Ruling and Ye, Yan and Chen, Zhibo},
  booktitle={Proceedings of the IEEE/CVF Conference on Computer Vision and Pattern Recognition},
  pages={6384--6394},
  year={2024}
}

@inproceedings{score,
  title={A kernelized Stein discrepancy for goodness-of-fit tests},
  author={Liu, Qiang and Lee, Jason and Jordan, Michael},
  booktitle={Int. Conf. Learn. Represent.},
  pages={276--284},
  year={2016},
  organization={PMLR}
}

@article{he2025videoscore2,
  title={Videoscore2: Think before you score in generative video evaluation},
  author={He, Xuan and Jiang, Dongfu and Nie, Ping and Liu, Minghao and Jiang, Zhengxuan and Su, Mingyi and Ma, Wentao and Lin, Junru and Ye, Chun and Lu, Yi and others},
  journal={arXiv preprint arXiv:2509.22799},
  year={2025}
}

@article{bai2025qwen3,
  title={Qwen3-vl technical report},
  author={Bai, Shuai and Cai, Yuxuan and Chen, Ruizhe and Chen, Keqin and Chen, Xionghui and Cheng, Zesen and Deng, Lianghao and Ding, Wei and Gao, Chang and Ge, Chunjiang and others},
  journal={arXiv preprint arXiv:2511.21631},
  year={2025}
}

@article{Qwen25VL,
  title={Qwen2.5-VL Technical Report},
  author={Bai, Shuai and Chen, Keqin and Liu, Xuejing and Wang, Jialin and Ge, Wenbin and Song, Sibo and Dang, Kai and Wang, Peng and Wang, Shijie and Tang, Jun and Zhong, Humen and Zhu, Yuanzhi and Yang, Mingkun and Li, Zhaohai and Wan, Jianqiang and Wang, Pengfei and Ding, Wei and Fu, Zheren and Xu, Yiheng and Ye, Jiabo and Zhang, Xi and Xie, Tianbao and Cheng, Zesen and Zhang, Hang and Yang, Zhibo and Xu, Haiyang and Lin, Junyang},
  journal={arXiv preprint arXiv:2502.13923},
  year={2025}
}

@article{zhu2025internvl3,
  title={Internvl3: Exploring advanced training and test-time recipes for open-source multimodal models},
  author={Zhu, Jinguo and Wang, Weiyun and Chen, Zhe and Liu, Zhaoyang and Ye, Shenglong and Gu, Lixin and Tian, Hao and Duan, Yuchen and Su, Weijie and Shao, Jie and others},
  journal={arXiv preprint arXiv:2504.10479},
  year={2025}
}

@article{qin2025ui,
  title={Ui-tars: Pioneering automated gui interaction with native agents},
  author={Qin, Yujia and Ye, Yining and Fang, Junjie and Wang, Haoming and Liang, Shihao and Tian, Shizuo and Zhang, Junda and Li, Jiahao and Li, Yunxin and Huang, Shijue and others},
  journal={arXiv preprint arXiv:2501.12326},
  year={2025}
}

@inproceedings{zhao2025swift,
  title={Swift: a scalable lightweight infrastructure for fine-tuning},
  author={Zhao, Yuze and Huang, Jintao and Hu, Jinghan and Wang, Xingjun and Mao, Yunlin and Zhang, Daoze and Jiang, Zeyinzi and Wu, Zhikai and Ai, Baole and Wang, Ang and others},
  booktitle={Proceedings of the AAAI Conference on Artificial Intelligence},
  volume={39},
  number={28},
  pages={29733--29735},
  year={2025}
}

@inproceedings{wu2022fast,
  title={Fast-vqa: Efficient end-to-end video quality assessment with fragment sampling},
  author={Wu, Haoning and Chen, Chaofeng and Hou, Jingwen and Liao, Liang and Wang, Annan and Sun, Wenxiu and Yan, Qiong and Lin, Weisi},
  booktitle={European conference on computer vision},
  pages={538--554},
  year={2022},
  organization={Springer}
}

@techreport{itu_p910_2023,
  title        = {{Subjective video quality assessment methods for multimedia applications}},
  author       = {{International Telecommunication Union}},
  institution  = {ITU-T},
  number       = {P.910},
  year         = {2023},
  note         = {Recommendation ITU-T P.910 (10/23)}
}

@article{wang2022internvideo,
  title={InternVideo: General Video Foundation Models via Generative and Discriminative Learning},
  author={Wang, Yi and Li, Kunchang and Li, Yizhuo and He, Yinan and Liu, Bei and Zhou, Zhen and Chen, Jiannan and Zhou, Yi and Wang, Ziyu and Xiao, Chen and others},
  journal={arXiv preprint arXiv:2212.03191},
  year={2022}
}

@article{breiman2001randomforest,
  title   = {Random Forests},
  author  = {Breiman, Leo},
  journal = {Machine Learning},
  volume  = {45},
  number  = {1},
  pages   = {5--32},
  year    = {2001}
}

@article{geurts2006extremely,
  title   = {Extremely Randomized Trees},
  author  = {Geurts, Pierre and Ernst, Damien and Wehenkel, Louis},
  journal = {Machine Learning},
  volume  = {63},
  number  = {1},
  pages   = {3--42},
  year    = {2006}
}

@article{hoerl1970ridge,
  title   = {Ridge Regression: Biased Estimation for Nonorthogonal Problems},
  author  = {Hoerl, Arthur E. and Kennard, Robert W.},
  journal = {Technometrics},
  volume  = {12},
  number  = {1},
  pages   = {55--67},
  year    = {1970}
}

@article{friedman2001greedy,
  title   = {Greedy Function Approximation: A Gradient Boosting Machine},
  author  = {Friedman, Jerome H.},
  journal = {The Annals of Statistics},
  volume  = {29},
  number  = {5},
  pages   = {1189--1232},
  year    = {2001}
}

@article{cover1967nearestneighbor,
  title   = {Nearest Neighbor Pattern Classification},
  author  = {Cover, Thomas and Hart, Peter},
  journal = {IEEE Transactions on Information Theory},
  volume  = {13},
  number  = {1},
  pages   = {21--27},
  year    = {1967}
}

@article{brooks2024video,
  title={Video generation models as world simulators},
  author={Brooks, Tim and Peebles, Bill and Holmes, Connor and DePue, Will and Guo, Yu and Jing, Li and Schnurr, David and Taylor, Joe and Luhman, Troy and Luhman, Eric and others},
  journal={OpenAI},
  year={2024}
}

@article{kling2024,
  title={Kling: A Powerful World Model for Video Generation},
  author={Kuaishou Technology},
  year={2024},
  url={https://kling.kuaishou.com/}
}

@article{agarwal2025cosmos,
  title={Cosmos world foundation model platform for physical ai},
  author={Agarwal, Niket and Ali, Arslan and Bala, Maciej and Balaji, Yogesh and Barker, Erik and Cai, Tiffany and Chattopadhyay, Prithvijit and Chen, Yongxin and Cui, Yin and Ding, Yifan and others},
  journal={arXiv preprint arXiv:2501.03575},
  year={2025}
}

@article{yang2024cogvideox,
  title={Cogvideox: Text-to-video diffusion models with an expert transformer},
  author={Yang, Zhuoyi and Teng, Jiayan and Zheng, Wendi and Ding, Ming and Huang, Shiyu and Xu, Jiazheng and Yang, Yuanming and Hong, Wenyi and Zhang, Xiaohan and Feng, Guanyu and others},
  journal={arXiv preprint arXiv:2408.06072},
  year={2024}
}

@article{zheng2024cami2v,
  title={Cami2v: Camera-controlled image-to-video diffusion model},
  author={Zheng, Guangcong and Li, Teng and Jiang, Rui and Lu, Yehao and Wu, Tao and Li, Xi},
  journal={arXiv preprint arXiv:2410.15957},
  year={2024}
}

@inproceedings{gu2025diffusion,
  title={Diffusion as shader: 3d-aware video diffusion for versatile video generation control},
  author={Gu, Zekai and Yan, Rui and Lu, Jiahao and Li, Peng and Dou, Zhiyang and Si, Chenyang and Dong, Zhen and Liu, Qifeng and Lin, Cheng and Liu, Ziwei and others},
  booktitle={Proceedings of the Special Interest Group on Computer Graphics and Interactive Techniques Conference Conference Papers},
  pages={1--12},
  year={2025}
}

@article{ex4d,
  title={EX-4D: EXtreme Viewpoint 4D Video Synthesis via Depth Watertight Mesh},
  author={Hu, Tao and Peng, Haoyang and Liu, Xiao and Ma, Yuewen},
  journal={arXiv preprint arXiv:2506.05554},
  year={2025}
}

@inproceedings{bai2025recammaster,
  title={Recammaster: Camera-controlled generative rendering from a single video},
  author={Bai, Jianhong and Xia, Menghan and Fu, Xiao and Wang, Xintao and Mu, Lianrui and Cao, Jinwen and Liu, Zuozhu and Hu, Haoji and Bai, Xiang and Wan, Pengfei and others},
  booktitle={Proceedings of the IEEE/CVF International Conference on Computer Vision},
  pages={14834--14844},
  year={2025}
}

@inproceedings{yu2025trajectorycrafter,
  title={Trajectorycrafter: Redirecting camera trajectory for monocular videos via diffusion models},
  author={Yu, Mark and Hu, Wenbo and Xing, Jinbo and Shan, Ying},
  booktitle={Proceedings of the IEEE/CVF international conference on computer vision},
  pages={100--111},
  year={2025}
}

@article{kong2024hunyuanvideo,
  title={Hunyuanvideo: A systematic framework for large video generative models},
  author={Kong, Weijie and Tian, Qi and Zhang, Zijian and Min, Rox and Dai, Zuozhuo and Zhou, Jin and Xiong, Jiangfeng and Li, Xin and Wu, Bo and Zhang, Jianwei and others},
  journal={arXiv preprint arXiv:2412.03603},
  year={2024}
}

@inproceedings{yu2023wonderjourney,
  title={Wonderjourney: Going from anywhere to everywhere},
  author={Yu, Hong-Xing and Duan, Haoyi and Hur, Junhwa and Sargent, Kyle and Rubinstein, Michael and Freeman, William T and Cole, Forrester and Sun, Deqing and Snavely, Noah and Wu, Jiajun and others},
  booktitle={Proceedings of the IEEE/CVF Conference on Computer Vision and Pattern Recognition},
  pages={6658--6667},
  year={2024}
}

@article{chen2025flexworld,
  title={FlexWorld: Progressively expanding 3D scenes for flexiable-view synthesis},
  author={Chen, Luxi and Zhou, Zihan and Zhao, Min and Wang, Yikai and Zhang, Ge and Huang, Wenhao and Sun, Hao and Wen, Ji-Rong and Li, Chongxuan},
  journal={arXiv preprint arXiv:2503.13265},
  year={2025}
}

@article{yu2024viewcrafter,
  title={ViewCrafter: Taming Video Diffusion Models for High-fidelity Novel View Synthesis},
  author={Yu, Wangbo and Xing, Jinbo and Yuan, Li and Hu, Wenbo and Li, Xiaoyu and Huang, Zhipeng and Gao, Xiangjun and Wong, Tien-Tsin and Shan, Ying and Tian, Yonghong},
  journal={IEEE Transactions on Pattern Analysis and Machine Intelligence},
  year={2025},
  publisher={IEEE}
}

@inproceedings{wang2024motionctrl,
  title={Motionctrl: A unified and flexible motion controller for video generation},
  author={Wang, Zhouxia and Yuan, Ziyang and Wang, Xintao and Li, Yaowei and Chen, Tianshui and Xia, Menghan and Luo, Ping and Shan, Ying},
  booktitle={ACM SIGGRAPH 2024 Conference Papers},
  pages={1--11},
  year={2024}
}

@inproceedings{duan2025worldscore,
  title={Worldscore: A unified evaluation benchmark for world generation},
  author={Duan, Haoyi and Yu, Hong-Xing and Chen, Sirui and Fei-Fei, Li and Wu, Jiajun},
  booktitle={Proceedings of the IEEE/CVF International Conference on Computer Vision},
  pages={27713--27724},
  year={2025}
}

@article{lu20254dworldbench,
  title={4DWorldBench: A Comprehensive Evaluation Framework for 3D/4D World Generation Models},
  author={Lu, Yiting and Luo, Wei and Tu, Peiyan and Li, Haoran and Zhu, Hanxin and Yu, Zihao and Wang, Xingrui and Chen, Xinyi and Peng, Xinge and Li, Xin and others},
  journal={arXiv preprint arXiv:2511.19836},
  year={2025}
}

@inproceedings{lu2024kvq,
  title={Kvq: Kwai video quality assessment for short-form videos},
  author={Lu, Yiting and Li, Xin and Pei, Yajing and Yuan, Kun and Xie, Qizhi and Qu, Yunpeng and Sun, Ming and Zhou, Chao and Chen, Zhibo},
  booktitle={Proceedings of the IEEE/CVF Conference on Computer Vision and Pattern Recognition},
  pages={25963--25973},
  year={2024}
}

@inproceedings{wu2023exploring,
  title={Exploring video quality assessment on user generated contents from aesthetic and technical perspectives},
  author={Wu, Haoning and Zhang, Erli and Liao, Liang and Chen, Chaofeng and Hou, Jingwen and Wang, Annan and Sun, Wenxiu and Yan, Qiong and Lin, Weisi},
  booktitle={Proceedings of the IEEE/CVF international conference on computer vision},
  pages={20144--20154},
  year={2023}
}

@inproceedings{huang2024vbench,
  title={Vbench: Comprehensive benchmark suite for video generative models},
  author={Huang, Ziqi and He, Yinan and Yu, Jiashuo and Zhang, Fan and Si, Chenyang and Jiang, Yuming and Zhang, Yuanhan and Wu, Tianxing and Jin, Qingyang and Chanpaisit, Nattapol and others},
  booktitle={Proceedings of the IEEE/CVF Conference on Computer Vision and Pattern Recognition},
  pages={21807--21818},
  year={2024}
}

@article{vbench2025,
  title={Vbench-2.0: Advancing video generation benchmark suite for intrinsic faithfulness},
  author={Zheng, Dian and Huang, Ziqi and Liu, Hongbo and Zou, Kai and He, Yinan and Zhang, Fan and Gu, Lulu and Zhang, Yuanhan and He, Jingwen and Zheng, Wei-Shi and others},
  journal={arXiv preprint arXiv:2503.21755},
  year={2025}
}

@inproceedings{meng2024phygenbench,
  title={Towards World Simulator: Crafting Physical Commonsense-Based Benchmark for Video Generation},
  author={Meng, Fanqing and Liao, Jiaqi and Tan, Xinyu and Lu, Quanfeng and Shao, Wenqi and Zhang, Kaipeng and Cheng, Yu and Li, Dianqi and Luo, Ping},
  booktitle={International Conference on Machine Learning},
  pages={43781--43806},
  year={2025},
  organization={PMLR}
}

@article{shang2026worldarena,
  title={WorldArena: A Unified Benchmark for Evaluating Perception and Functional Utility of Embodied World Models},
  author={Shang, Yu and Li, Zhuohang and Ma, Yiding and Su, Weikang and Jin, Xin and Wang, Ziyou and Jin, Lei and Zhang, Xin and Tang, Yinzhou and Su, Haisheng and others},
  journal={arXiv preprint arXiv:2602.08971},
  year={2026}
}

@inproceedings{hessel2021clipscore,
  title={Clipscore: A reference-free evaluation metric for image captioning},
  author={Hessel, Jack and Holtzman, Ari and Forbes, Maxwell and Le Bras, Ronan and Choi, Yejin},
  booktitle={Proceedings of the 2021 conference on empirical methods in natural language processing},
  pages={7514--7528},
  year={2021}
}

@inproceedings{zhang2025q,
  title={Q-Bench-Video: Benchmark the Video Quality Understanding of LMMs},
  author={Zhang, Zicheng and Jia, Ziheng and Wu, Haoning and Li, Chunyi and Chen, Zijian and Zhou, Yingjie and Sun, Wei and Liu, Xiaohong and Min, Xiongkuo and Lin, Weisi and others},
  booktitle={Proceedings of the Computer Vision and Pattern Recognition Conference},
  pages={3229--3239},
  year={2025}
}

@inproceedings{guan2025internvqa,
  title={Internvqa: Advancing compressed video quality assessment with distilling large foundation model},
  author={Guan, Fengbin and Yu, Zihao and Lu, Yiting and Li, Xin and Chen, Zhibo},
  booktitle={2025 IEEE International Symposium on Circuits and Systems (ISCAS)},
  pages={1--5},
  year={2025},
  organization={IEEE}
}
}


\end{document}